\documentclass[10pt,twocolumn,letterpaper]{article}

\usepackage[pagenumbers]{cvpr} 

\usepackage[dvipsnames]{xcolor}
\usepackage{graphicx}
\usepackage{multirow}
\usepackage{tabularx}
\usepackage{wrapfig}
\usepackage[acronym,toc]{glossaries}
\usepackage[flushleft]{threeparttable}

\definecolor{cvprblue}{rgb}{0.21,0.49,0.74}
\usepackage[pagebackref,breaklinks,colorlinks,citecolor=cvprblue]{hyperref}

\def\confName{CVPR}
\def\confYear{2024}

\title{Audio-Visual Compound Expression Recognition Method based on Late Modality Fusion and Rule-based Decision}

\author{Elena Ryumina\\
St. Petersburg Federal Research Center \\ of the Russian Academy of Sciences\\
St. Petersburg, Russia\\
{\tt\small ryumina.e@iias.spb.su}
\and
Maxim Markitantov\\
St. Petersburg Federal Research Center \\ of the Russian Academy of Sciences\\
St. Petersburg, Russia\\
{\tt\small markitantov.m@iias.spb.su}
\and
Dmitry Ryumin\\
St. Petersburg Federal Research Center \\ of the Russian Academy of Sciences\\
St. Petersburg, Russia\\
{\tt\small ryumin.d@iias.spb.su}
\and
Heysem Kaya\\
Department of Information and Computing Sciences\\
Utrecht University, The Netherlands\\
{\tt\small h.kaya@uu.nl}
\and
Alexey Karpov\\
St. Petersburg Federal Research Center \\ of the Russian Academy of Sciences\\
St. Petersburg, Russia\\
{\tt\small karpov@iias.spb.su}
}

\begin{document}
\maketitle

\newacronym{CER}{CER}{Compound Expression Recognition}
\newacronym{CE}{CE}{Compound Expression}
\newacronym{ABAW}{ABAW}{Affective Behavior Analysis in-the-Wild}
\newacronym{AFEW}{AFEW}{Acted Facial Expressions in The Wild}

\newacronym{VS}{VS}{Static visual model}
\newacronym{VD}{VD}{Dynamic visual model}
\newacronym{FCL}{FCL}{Fully Connected Layer}
\newacronym{FPS}{FPS}{Frame Per Second}
\newacronym{LSTM}{LSTM}{Long Short-Term Memory}
\newacronym{VAD}{VAD}{Voice Activity Detection}

\newacronym{Ne}{Ne}{Neutral}
\newacronym{An}{An}{Anger}
\newacronym{Di}{Di}{Disgust}
\newacronym{Fe}{Fe}{Fear}
\newacronym{Ha}{Ha}{Happiness}
\newacronym{Sa}{Sa}{Sadness}
\newacronym{Su}{Su}{Surprise} 

\begin{abstract}
This paper presents the results of the SUN team for the \acrfull{CE} Recognition Challenge of the 6th ABAW Competition. We propose a novel audio-visual method for compound expression recognition. Our method relies on emotion recognition models that fuse modalities at the emotion probability level, while decisions regarding the prediction of compound expressions are based on predefined rules. Notably, our method does not use any training data specific to the target task. Thus, the problem is a zero-shot classification task. The method is evaluated in multi-corpus training and cross-corpus validation setups. Using our proposed method is achieved an F1-score value equals to 22.01\% on the C-EXPR-DB test subset. Our findings from the challenge demonstrate that the proposed method can potentially form a basis for developing intelligent tools for annotating audio-visual data in the context of human's basic and compound emotions.
\end{abstract}
  
\section{Introduction}
\label{sec:intro}

The \gls{CER} as a part of affective computing is a novel task in intelligent human-computer interaction and multimodal user interfaces. It entails the automated identification of compound emotional states in individuals, which may include combinations of two or more basic emotions such as: Fear, Happiness, Sadness, Anger, Surprise, and Disgust.

Over the last two decades, research efforts in the field of automatic expression analysis have predominantly focused on identifying six basic emotions~\cite{RYUMINA2022435,kollias2023multi}. However, these methods fail to fully capture the complexity of everyday emotional expressions. Individuals often exhibit \glspl{CE}, such as Fearfully Surprised, Happily Surprised, Sadly Surprised, Disgustedly Surprised, Angrily Surprised, Sadly Fearful, Sadly Angry, which are combinations of basic emotions. This \glspl{CE} underscores the necessity for more comprehensive models capable of capturing the subtleties inherent in human's emotional expressions.

Existing methods for \gls{CER} predominantly focus on the visual modality~\cite{liu2022mafw,sun2023mae}. These methods use both dynamic~\cite{liu2022mafw,sun2023mae,sun2024hicmae} and static~\cite{li2017reliable,liu2022mafw,kollias2023multi} deep models, relying on facial action units~\cite{li2017reliable,li2023compound,kollias2023multi}. The audio models used in the first two audio-visual methods are based on spectrograms~\cite{sun2024hicmae,kollias2023multi}. However, to train a model for \gls{CER}, it is necessary to have relevant data comprising balanced samples for each class, collected under uncontrolled conditions, containing multimodal data, and large enough to train deep neural network models. Nevertheless, challenges in annotating \glspl{CE}~\cite{kollias2023multi} contribute to the scarcity of such corpora. An exception is the C-EXPR-DB corpus~\cite{kollias2023multi,kollias2023abaw2,kollias2023abaw,kollias2022abaw}, a part of which is presented as a Test subset in the 6th Workshop and Competition on \gls{ABAW}~\cite{kollias20246th}. Another corpus along with the \glspl{CE} is a Multi-modal compound Affective database for facial expression recognition in the Wild~(MAFW)~\cite{liu2022mafw}, which has limited access.

In this paper, we present a novel method for audio-visual \gls{CER} as a part of the 6th \gls{ABAW} \gls{CE} Recognition Challenge. Our method does not utilize the \gls{CER} challenge data as training data; rather, it includes models trained for basic emotion recognition. Decisions regarding predicted \glspl{CE} are determined by the proposed rule-based method. It enables us to address the problem of insufficient publicly available data with \glspl{CE}. Moreover, \glspl{CE} comprise various pair combinations of basic emotions, rendering the proposed method particularly valuable to the research community. The versatility of the proposed method allows for use as a software tool for annotating new emotionally colored data.

In summary, our main contributions are as follows:
\begin{itemize}
\item We introduce a novel audio-visual \gls{CER} method based on basic emotion recognition and analysis of emotion probability distribution through multimodal fusion.
\item We present a method for audio-visual emotion recognition based on both multi-corpus and cross-corpus research.
\item We propose a rule-based decision-making method for \gls{CER} that explains which modality is responsible for predicting specific \glspl{CE}.
\item We provide new baseline performance measures for the recognition task of the seven basic emotions on the Validation subsets of the AffWild2 and \gls{AFEW} corpora.
\end{itemize}

\begin{figure*}
  \centering
   \includegraphics[width=0.95\linewidth]{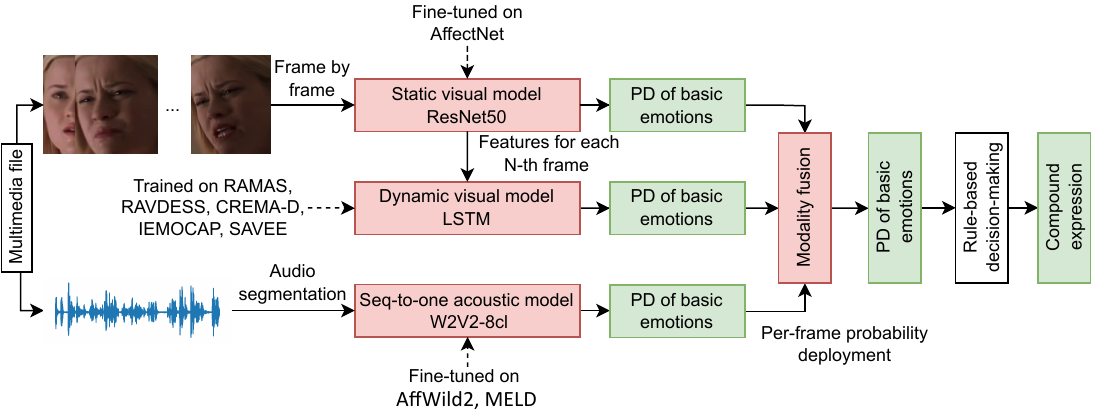}
   \caption{Pipeline of the proposed audio-visual \gls{CER} method. PD refers to probability distribution.}
   \label{fig:pipeline}
\end{figure*}

\section{Proposed Method}
\label{sec:method}
A pipeline of the proposed audio-visual \gls{CER} method is shown in Figure~\ref{fig:pipeline}. The method accepts a multimedia file as an input. Then, it performs the necessary pre-processing for each modality, including face region detection and audio data extraction. The pre-processed data are used as an input for three models, which predict the probability distribution of recognized emotions. A hierarchical probability weighting is then applied. The final weighted probabilities are utilized for rule-based \glspl{CE} prediction.

\subsection{Video Models}
We use the RetinaFace model~\cite{deng2020retinaface}, as implemented by\footnote{https://github.com/hhj1897/face\_detection}, for face region detection. However, relying only on a single detector is insufficient; it is necessary to perform post-processing on the detected face regions, including determining the target person, removing erroneously detected face regions, etc. Although the \gls{CE} Recognition Challenge invites recognizing \glspl{CE} frame by frame, using only a static model is also inappropriate. For example, transitioning from one expression to another may involve a neutral state or other intermediate states. Therefore, our method integrates the recognition of \glspl{CE} using both static and dynamic visual models.

\textbf{\gls{VS}.} As a static model for affective state recognition (comprising six basic emotions and a neutral state), we utilize the ResNet50 model~\cite{he2016deep} pre-trained on face recognition\footnote{https://github.com/rcmalli/keras-vggface}. The model extracts discriminative features from faces that are useful for transfer learning in our task. We initialize the model with pre-trained weights and subsequently fine-tune it to recognize affective states without freezing its layers. For feature extraction and classification, we extend the model by adding two \glspl{FCL} comprising 512 and 7 neurons, respectively. In the proposed method, we use a static model to detect affective states in each frame and extract features from every $N$ frame (where $N$ is the frame sampling step). We utilize these features as inputs for the dynamic model.

\textbf{\gls{VD}.} The model designed for analyzing dynamically changing affective states operates on 2-second segments or 10 frames. To produce 10 frames within two seconds, the \gls{FPS} rate is reduced to five \gls{FPS}. The model proposed comprises two \gls{LSTM} layers, with 512 and 256 neurons, respectively. It also includes a classification layer consisting of 7 neurons.

To enhance the generalizability of the video models, several augmentation techniques are applied, namely MixUp~\cite{zhang2017mixup} and Label Smoothing~\cite{lukasik2020does}. These techniques help to reduce the confidence level of the models in their basic emotion predictions, thereby enabling them to identify multiple emotions with varying degrees of certainty in the frames. All the models are trained using the Adam optimizer with a learning rate of 1e-4 for 30 epochs and the Cosine Annealing Cold Restart Learning Scheduler~\cite{loshchilov2016sgdr} with five rate restart cycles.

\subsection{Audio Model}
Before training an audio model, in addition to extracting audio signals from multimedia files, we detect voice activity. Two approaches are used for this purpose, depending on the corpus used. The first one employs an audio-based \gls{VAD}\footnote{https://github.com/snakers4/silero-vad}. The second one relies on the video modality, analyzing video data frame by frame. We extract facial landmarks using MediaPipe~\cite{lugaresi2019mediapipe}, then mouth landmarks are identified and the region of interest is extracted. It is used to determine whether the target speaker's mouth is open or closed. We employ this method due to the specificity of the training acoustic data, which may include background noises, making it challenging to identify the target speaker. Then, 4-second segments with a step of two seconds are formed on the detected segments of voice activity. In addition, to obtain the target label of a window, we compute the most frequent frame-wise label.

\textbf{Sequence-to-One acoustic model}. We proposed two slightly different models. The backbone of both models is based on the pre-trained emotional model Wav2vec2~\cite{wagner2023dawn}. This model was pre-trained using the regression emotion dimensions (arousal, valence, and dominance) from the MSP-Podcast corpus~\cite{lotfian2017building}. On top of the model, we stack two transformer layers with self-attention mechanisms, each with 32 and 16 heads. After the last transformer layer, we aggregate the information along the time axis and apply a \gls{FCL} with seven or eight neurons, depending on the number of classes. We fine-tune all the layers from the top to the last two (W2V2-7cl) or four (W2V2-8cl) encoding layers of the backbone model for models with seven and eight neurons, respectively.

Similar to the video model, Label Smoothing~\cite{lukasik2020does} is also used for the audio model to reduce the confidence of the model. The remaining training parameters are identical to those of the video model.

\subsection{Modality Fusion}
The proposed modality fusion method uses three models to represent emotion probability distributions. Each model exhibits varying prediction confidences for different emotions. Therefore, we employ a hierarchical probability weighting before predicting \glspl{CE}. The importance of models and probabilities is considered in the first weighting. The weight matrix $W$ is generated using the Dirichlet distribution:
\begin{equation}
W = \begin{bmatrix}
        w_{11} & w_{12} & \cdots & w_{1C} \\
        \vdots & \vdots & \ddots & \vdots \\
        w_{M1} & w_{M2} & \cdots & w_{MC}
    \end{bmatrix}
\end{equation}
where $W \in \mathbb{R}^{M \times C}$,  $M$ is the number of models and $C$ is the number of emotional classes. The weight matrix is generated such that the sum of the weights for the three models of each class equals one. After the first weighting, new probabilities for each model are calculated using the following formula:
\begin{equation}
  \bar{P}_M = P_{M} \times w_{M},
  \label{eq:first_weighting}
\end{equation}
where $P_{M}=[p_{M1}, p_{M2}, ... , p_{MC}]$ is the probability vector for the model $M$, $w_{M}=[w_{M1}, w_{M2}, ... , w_{MC}]$ is the weight vector for the model $M$. In the second weighting, only the importance of the models is taken into account. A weight vector $V$ of size $M$ is generated with one value for each model. The vector values are generated in the range of [0.01; 0.5] with an increment of 0.005. A final probability vector $\hat{P}$ is obtained using the following formula:
\begin{equation}
  \hat{P} = \sum_{i=1}^{M} \bar{P}_{i} \times v_{i},
  \label{eq:final_proba}
\end{equation}
where $v_1 \in V$. $W$ and $V$ are weights that remain consistent across all test samples. This hierarchical weighting enhances performance measures for basic emotion recognition by considering the contribution of each model. The final probability vector is then utilized for \gls{CER}.

\subsection{Rule-based Decision-Making Method}
We utilize two rules to make decisions regarding the predicted \glspl{CE}. The first rule (Rule 1) allow for certain emotion predictions and is based on masking probabilities that fall below the minimum threshold (1/7) for emotion prediction. This rule is exclusively applied to the outputs of Dirichlet-based fusion method, e.g. $\bar{P}$ probability vectors. The probability vector is updated according to the following condition:
\begin{equation}
\label{eq:r_value}
    \bar{P}  = \left\{
            \begin{array}{ll}
              0,  &  if\,\, \bar{p}^E < 1/7 \\
              \bar{p}^E,  & else\\
            \end{array},
          \right.
\end{equation}
where $\bar{p}^E$ is the probability of the emotion $E$. For the Rule 1, according to the probabilities of the basic emotions \gls{Ne}, \gls{An}, \gls{Di}, \gls{Fe}, \gls{Ha}, \gls{Sa}, and \gls{Su}, the probability value for the Fearfully Surprised $\bar{cp}^{FeSu}$ is calculated using the following formula:
\begin{equation}
  \bar{cp}^{FeSu} = \bar{p}^{Fe} + \bar{p}^{Su},
  \label{eq:first_rule}
\end{equation}
where $\bar{p}^{Fe}$ and $\bar{p}^{Su}$ are the probability values for \gls{Fe} and \gls{Su} emotions, respectively. The probabilities for the remaining \glspl{CE} are calculated similarly. The predicted class is the class with the maximal probability.

\begin{table}[!h]
  \centering
  \resizebox{\columnwidth}{!}{%
  \begin{tabular}{lllll}
    \toprule
    \gls{CE} class & E\textsubscript{1}  & CW\textsubscript{1} & E\textsubscript{2}  & CW\textsubscript{2}\\
    \midrule
    Fearfully Surprised&Fear& 5/7 &Surprise& 2/7\\
    Happily Surprised&Happiness& 6/8 &Surprise& 2/8\\
    Sadly Surprised&Sadness&4/6&Surprise& 2/6\\
    Disgustedly Surprised&Disgust&6/8&Surprise& 2/8\\
    Angrily Surprised&Anger&5/7&Surprise& 2/7\\
    Sadly Fearful&Sadness&4/9&Fear& 5/9\\
    Sadly Angry&Sadness&4/9&Anger& 5/9\\
    \bottomrule
  \end{tabular}}
  \caption{Weights of basic emotions used for \gls{CER}. E\textsubscript{1} and E\textsubscript{2} refer to the first and second emotion in a pair, CW\textsubscript{1} and CW\textsubscript{2} to the weights for the first and second emotion in a pair.}
  \label{tab:weights_basic_emotion}
\end{table}

\begin{table*}[t]
\centering
\resizebox{\textwidth}{!}{
\begin{tabular}{@{}lccccccccc@{}}
\toprule
\multirow{3}{*}{ID}&\multirow{3}{*}{Model} & \multirow{3}{*}{Train corpus} & \multicolumn{7}{c}{Test corpus} \\
&& & \multicolumn{2}{c}{AffWild2 (7cl)} & \multicolumn{2}{c}{AFEW (7cl)} & \multicolumn{3}{c}{C-EXPR-DB (7cl), F1} \\
&& & F1 & UAR& F1 & UAR & Rule 1 & Rule 2 & W/o rules\\
\midrule
1& Static visual model, ResNet50& AffectNet & 34.71 & 40.33 & 42.83 & 43.75 & \multicolumn{1}{c}{--} & -- & -- \\
\multirow{2}{*}{2}& \multirow{2}{*}{Dynamic visual model, LSTM} & RAMAS, RAVDESS, CREMA-D, & \multirow{2}{*}{39.71} & \multirow{2}{*}{42.44} & \multirow{2}{*}{41.82} & \multirow{2}{*}{43.59} & \multirow{2}{*}{--} & \multirow{2}{*}{--} & \multirow{2}{*}{--} \\
&  & IEMOCAP, SAVEE & &  &  &  &  &  &  \\
3& Model ID 1 \& 2 (Dirichlet-based weighing) & -- & 42.10 & 45.64 & 44.09 & 43.98 & -- & -- & --\\
4& Model ID 1 \& 2  (hierarchical weighing) & -- & 42.27 & 46.19 & 43.79 & 44.84 & -- & 19.44 & 20.98 \\
5& Seq-to-one acoustic model, W2V2-7cl & AffWild2, MELD & 31.11 & 30.76 & 22.88 & 26.81 & -- & -- & -- \\
6& Seq-to-one acoustic model, W2V2-8cl &  AffWild2, MELD & 31.56 & 33.64 & 22.82 & 24.53 & -- & -- & -- \\
7& Model ID 1 \& 2 \& 5 (Dirichlet-based weighing) & -- & 44.51 & 49.51 & 43.86 & 44.66 & -- & -- & -- \\
8& Model ID 1 \& 2 \& 5 (hierarchical weighing) & -- & 42.77 & 49.98 & 43.09 & 43.55 & -- & 17.56 & -- \\
9& Model ID 1 \& 2 \& 6 (Dirichlet-based weighing) & -- & \textbf{46.79} & \textbf{51.79} & \textbf{44.81} & \textbf{45.66} & 21.14 & \textbf{22.01} & -- \\
10&Model ID 1 \& 2 \& 6 (hierarchical weighing) & -- & 39.36 & 46.60 & 35.76 & 37.70& -- & -- & -- \\
\bottomrule
\end{tabular}
}
\caption{Experimental results of basic emotions and CEs recognition.}
  \label{tab:Results}
\end{table*}

The second rule (Rule 2) is based on weighting the frequency of emotions occurring in \glspl{CE}. In the \gls{CE} Recognition Challenge, we aim to develop a method for recognizing seven \glspl{CE}: Fearfully Surprised, Happily Surprised, Sadly Surprised, Disgustedly Surprised, Angrily Surprised, Sadly Fearful, Sadly Angry. The occurrence frequency of each emotion differs among these pairs. For example, the emotion of Surprise is more frequent than the others; therefore, this emotion does not allow you to distinguish \glspl{CE}. The use of emotion weights can enhance the importance of the probability of less represented emotions. The weight vectors, $CW_1$ and $CW_2$, determine the weights of the first and second emotions in pairs of \glspl{CE}. The weight vectors, whose paired values sum to one, are experimentally determined for our method and are shown in Table~\ref{tab:weights_basic_emotion}.

For the Rule 2, the probability value for the Fearfully Surprised $\bar{cp}^{FeSu}$ is calculated based on the established weights (see Table~\ref{tab:weights_basic_emotion}) and on the probabilities of the basic emotions using the formula:
\begin{equation}
  \bar{cp}^{FeSu} = \bar{p}^{Fe} \times cw_1^{FeSu} + \bar{p}^{Su} \times cw_2^{FeSu},
  \label{eq:second_rule}
\end{equation}
where $cw_1^{FeSu} \in CW_1$ and $cw_2^{FeSu} \in CW_2$. This rule is applied to the outputs of both Dirichlet-based ($\bar{P}$) and the hierarchical modality fusion ($\hat{P}$) methods.
 
\section{Experiments}

\subsection{Research corpora}
We propose the audio-visual \gls{CER} method based on the models designed for recognizing six basic emotions and a neutral state. 

We use several corpora for training, validating and testing the developed emotion recognition models. To train a static video model, we use the AffectNet corpus~\cite{mollahosseini2017affectnet}. This corpus comprises an extensive collection of static facial images displaying spontaneous emotions. We use the RAMAS~\cite{perepelkina2018ramas}, RAVDESS~\cite{livingstone2018ryerson}, CREMA-D~\cite{cao2014crema}, IEMOCAP~\cite{busso2008iemocap} and SAVEE~\cite{haq2008audio} corpora to train the dynamic visual model. In contrast to AffectNet, these corpora were collected in office conditions, but contain dynamically changing expressions. Therefore, the annotation quality is considered reliable, the facial images are noiseless, and the multi-corpus training provides the model with a high generalization ability to new data.

To train the audio model, we conduct a multi-corpus training using the AffWild2~\cite{kollias2021analysing,kollias2020analysing,kollias2021distribution,kollias2021affect,kollias2019expression,kollias2019face,kollias2019deep,zafeiriou2017aff} and MELD~\cite{poria2018meld} corpora. These corpora comprise recordings collected in uncontrolled conditions and include various paralinguistic elements in speech (such as laughter, shouting, etc.), making the data more relevant to real-world scenarios in comparison to the aforementioned corpora.

Validation of the audio and video models, as well as optimization of the modality fusion weights, is conducted on the Validation subset of the AffWild2 corpus (with corpus annotations dated 2024). To ensure that we do not overfit the models and fusion weights for each corpus, we use the \gls{AFEW} Validation subset~\cite{dhall2019emotiw} as an additional validation corpus. Finally, the \gls{CER} method is tested on the non-annotated Test subset of the C-EXPR-DB corpus.

\subsection{Experimental Results}

\begin{figure}[!ht]
     \centering
     \begin{subfigure}[c]{\linewidth}
         \includegraphics[width=\linewidth]{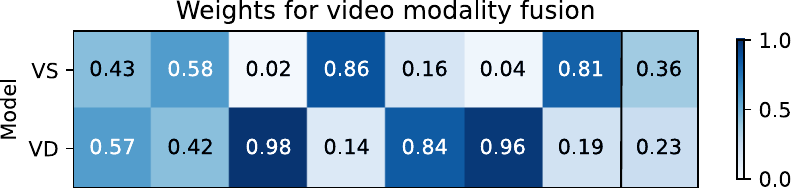}
     \end{subfigure}
     \begin{subfigure}[c]{\linewidth}
         \includegraphics[width=\linewidth]{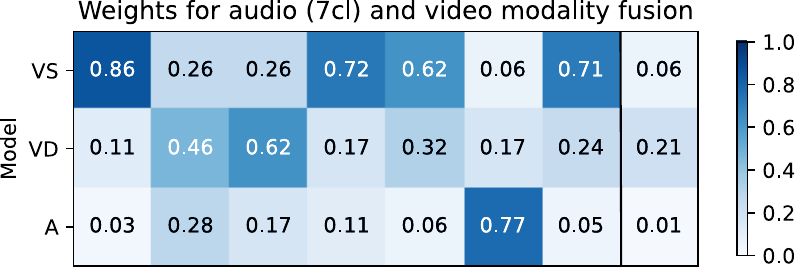}
     \end{subfigure}
     \begin{subfigure}[c]{\linewidth}
         \includegraphics[width=\linewidth]{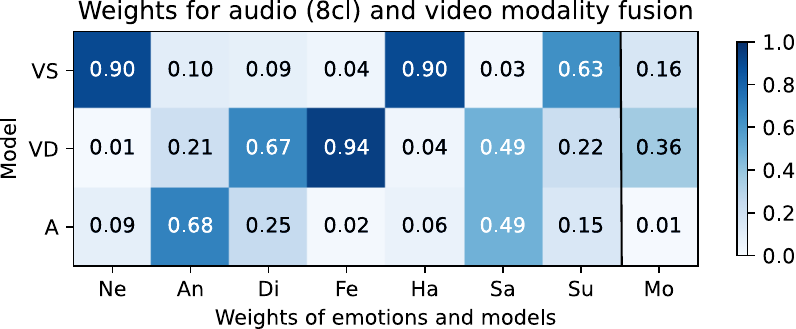}
     \end{subfigure}
    \caption{Weights for different modality fusion. VS, VD, and A refer to static visual, dynamic visual, and acoustic models, respectively. \gls{Ne}, \gls{An}, \gls{Di}, \gls{Fe}, \gls{Ha}, \gls{Sa}, and \gls{Su} refer to the weights of seven emotions used for Dirichlet-based weighting, Mo to the weights of models used for hierarchical weighting.}
    \label{fig:w1}
\end{figure}

\begin{figure*}[!ht]
     \centering
     \begin{subfigure}[l]{\linewidth}
     \includegraphics[width=\linewidth]{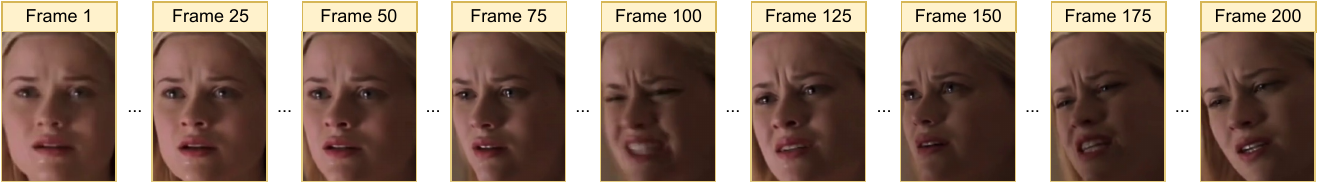}
     \end{subfigure}
     \begin{subfigure}[l]{\linewidth}
         \includegraphics[width=\linewidth]{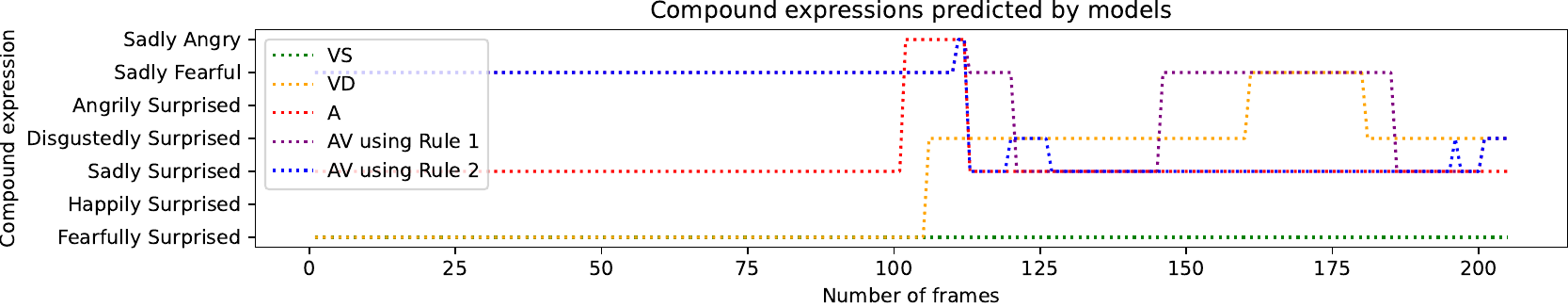}
     \end{subfigure}
    \caption{An example of \glspl{CE} prediction using video from the C-EXPR-DB corpus.}
    \label{fig:pred}
\end{figure*}

The experimental results are presented in Table~\ref{tab:Results}. Depending on the corpus, the visual models show different performance; the dynamic model outperforms the static model for AffWild2, and vice versa for \gls{AFEW}. When the two visual models are combined, the hierarchical weighting fusion outperforms the Dirichlet-based weighting. Therefore, in the first submission to the \gls{CE} Recognition Challenge, we use Model ID 4 in two versions: with and without the Rule 2 to predict \gls{CE}. Using Rule 2 reduces the performance measure of \gls{CER} by 1.54\% compared to using the rule-free method. This suggests that increasing the importance of the less represented emotions in \gls{CE} pairs does not improve \gls{CER} performance.

We then evaluate the performance of the audio models on two corpora, AffWild2 and \gls{AFEW}. It is impossible to say which model is more efficient. However, since these models show lower performance than the video models, we do not test only the audio models on C-EXPR-DB. Finally, we fuse the audio and video models by comparing various weighting fusions. The results show that the hierarchical weighting fusion outperforms the Dirichlet-based fusion. However, for comparison with previous models tested on C-EXPR-DB, we submit Model ID 8 with the Rule 2 for \gls{CE} prediction. Additionally, we evaluate Model ID 9 based on Dirichlet fusion with both rules. Combining the visual models with the audio model results in a 2.57\% improvement in F1. However, employing hierarchical weighting for model fusion shows a performance decrease of 4.45\% compared to Dirichlet-based weighting. Additionally, Rule 1 exhibits a 0.87\% lower performance than Rule 2. 

To understand the significance of each model in the final \gls{CE} predictions, the fusion model weights are presented in Figure~\ref{fig:w1}. The analysis of the fusion weights involving only the two visual models indicates a preference towards the dynamic model for predicting \gls{Di}, \gls{Ha}, and \gls{Sa} expressions, while favoring the static model for the predicting \gls{Fe} and \gls{Su} expressions. The hierarchical weighting reduces the contribution of the dynamic model. This weight distribution suggests that considering the \gls{CE} weighting (see Table~\ref{tab:weights_basic_emotion}), the method bases its decision on the dynamic model for predicting, for example, the Disgustedly Surprised and Happily Surprised classes, whereas it relies on the static model for predicting classes like Fearfully Surprised and Sadly Fearful.

The acoustic model trained on seven classes contributes less to the final prediction than the model trained on eight classes. For example, the first model holds greater in predicting only the \gls{Sa} emotion, whereas the second model demonstrates a higher contribution in predicting \gls{An} and competes with the dynamic video model in predicting \gls{Sa}. Nevertheless, in both cases, the hierarchical weighting leads to a complete disregard of the acoustic model, consequently leading to a decrease in emotion recognition performance. Thus, when combining three models, using hierarchical weighting of emotion probability distribution proves ineffective. The latter fusion (see Figure~\ref{fig:w1}, bottom sub-figure) demonstrates that the method bases its decision on the acoustic model when predicting, for example, the Angrily Surprised and Sadly Angry classes, while relying on the static model to predict the Happily Surprised class. The contribution of the dynamic model is considered when predicting other \glspl{CE}.

We also show an example of \gls{CER} in Figure~\ref{fig:pred}. From the frames depicted, it is clear that the woman is experiencing negative \glspl{CE}; none of the models make a mistake in predicting Happily Surprised. All models, except the \gls{VS} model, detect a change in expression at frame 100. Moreover, it is the audio model that recognizes the \gls{An} emotion, which also influences the correct prediction of Sadly Angry by the audio-visual models. In general, according to the predictions of all models, the frames presented contain \glspl{CE} such as: Fearfully Surprised, Sadly Surprised, Disgustedly Surprised, Sadly Fearful, and Sadly Angry. Subjectively, we have a tendency to assume that the video represents \glspl{CE} such as: Sadly Fearful, Sadly Angry, Angrily Disgusted (this \gls{CE} is not in the target classes). Thus, the most accurate predictor of \glspl{CE} in this video is the audio-visual model with the Rule 1 (a purple line).

Therefore, we have developed a method comprising three models, each assigned responsibility for predicting its respective class during \gls{CER}.

\section{Conclusions}

In this paper, we propose a novel audio-visual method for \gls{CER}. The method integrates three models, including the static and dynamic visual models, as well as the audio model. Each model predicts the emotion probabilities for six basic emotions and the neutral state. The emotional probabilities are then weighted utilizing the Dirichlet distribution. Finally, two rules are applied to determine \gls{CE}. Additionally, We provide new baselines for recognizing seven emotions on the Validation subsets of the AffWild2 and \gls{AFEW} corpora.

The experimental results demonstrate that each model is responsible for predicting specific \glspl{CE}. For example, the audio model is responsible for predicting the Angry Surprised and Sadly Angry, the static visual model is responsible for predicting the Happily Surprised class, and the dynamic visual model well predicts other \glspl{CE}. The results obtained show that the proposed method can potentially lead to intelligent software tools for fast annotation of data containing both basic and compound emotional expressions.

\section{Acknowledgements}
\label{sec:acknowledgments}
This study was supported by the Russian Science Foundation (project No. 22-11-00321).

{
    \small
    \bibliographystyle{ieeenat_fullname}
    \bibliography{main}
}

\end{document}